\documentclass[11pt,a4paper]{article}
\usepackage[hyperref]{emnlp2020}
\usepackage{times}
\usepackage{latexsym}

\usepackage{microtype}
\usepackage{amsmath}
\usepackage{amssymb}

\usepackage{graphicx}
\usepackage{comment}
\usepackage{array}
\usepackage{mathtools}
\usepackage[linguistics]{forest}
\usepackage{xcolor}
\usepackage{comment}

\aclfinalcopy % Uncomment this line for the final submission
\makeatletter
\newcommand{\printfnsymbol}[1]{%
  \textsuperscript{\@fnsymbol{#1}}%
}
\makeatother

\title{Introducing Syntactic Structures into Target Opinion Word Extraction with Deep Learning}

\author{Amir Pouran Ben Veyseh\textsuperscript{\rm 1}\thanks{\text{ } Equal contribution.}, \text{ } Nasim Nouri\printfnsymbol{1}, Franck Dernoncourt\textsuperscript{\rm 2},\\
{\bf Dejing Dou}\textsuperscript{\rm 1} and {\bf Thien Huu Nguyen}\textsuperscript{\rm 1,3} \\
\textsuperscript{\rm 1} Department of Computer and Information Science, University of Oregon,
\\Eugene, OR 97403, USA\\
\textsuperscript{\rm 2} Adobe Research, San Jose, CA, USA\\
\textsuperscript{\rm 3} VinAI Research, Vietnam\\
  \texttt{\{apouranb,dou,thien\}@cs.uoregon.edu}, \\ {\tt nasim.nourii@gmail.com}, {\tt dernonco@adobe.com}
}

\date{}

\begin{document}
\maketitle
\begin{abstract}
Targeted opinion word extraction (TOWE) is a sub-task of aspect based sentiment analysis (ABSA) which aims to find the opinion words for a given aspect-term in a sentence. Despite their success for TOWE, the current deep learning models fail to exploit the syntactic information of the sentences that have been proved to be useful for TOWE in the prior research. In this work, we propose to incorporate the syntactic structures of the sentences into the deep learning models for TOWE, leveraging the syntax-based opinion possibility scores and the syntactic connections between the words. We also introduce a novel regularization technique to improve the performance of the deep learning models based on the representation distinctions between the words in TOWE. The proposed model is extensively analyzed and achieves the state-of-the-art performance on four benchmark datasets.

%Deep learning models have been shown to achieve the state-of-the-art performance for TOWE in the recent studies.

%While previous feature-based models have shown syntactical structure (i.e., dependency tree) is useful for this task, recent deep neural nets ignore this information in their model. To address this limitation, in this paper, we propose a new approach which incorporates syntactical structure (i.e., dependency tree) into deep neural nets. More specifically, our model employs the dependency tree to capture the relative importance of the words to the aspect-term and to encode the connections between words. Our extensive experiments on four benchmark datasets prove the superiority of the proposed model, leading to new state-of-the-art results on all datasets. Moreover, detailed analysis shows the effectiveness of the components of the proposed model. 
\end{abstract}

\section{Introduction}

Targeted Opinion Word Extraction (TOWE) is an important task in aspect based sentiment analysis (ABSA) of sentiment analysis (SA). Given a target word (also called aspect term) in the input sentence, the goal of TOWE is to identify the words in the sentence (called the target-oriented opinion words) that help to express the attitude of the author toward the aspect represented by the target word. For instance, as a running example, in the sentence ``\textit{All warranties honored by XYZ (what I thought was a reputable company) are disappointing.}", ``\textit{disappointing}" is the opinion word for the target word ``\textit{warranties}" while the opinion words for the target word ``\textit{company}" would involve ``\textit{reputable}''. Among others, TOWE finds its applications in target-oriented sentiment analysis \citep{tang2015effective,xue2018aspect,Veyseh:20e} and opinion summarization \citep{wu2020latent}.

%As the opinion words might provide useful information to explain and/or improve the sentiment prediction of the ABSA systems, TOWE can be applied in different problems, including target-oriented sentiment analysis \citep{tang2015effective,wang2016attention,xue2018aspect,wu2020latent} and pair-wise opinion summarization \citep{hu2004mining,zhuang2006movie,wu2020latent}.

%A notable problem is that although the related tasks of TOWE has been extensively explored in the past, there have been only a few work to explicitly consider the TOWE problem in the literature \citep{fan2019target}. In particular, the most related task of TOWE is opinion word extraction (OWE) that aims to locate the terms used to express attitude explicitly in the sentence \citep{htay2013extracting, shamshurin2012extracting}. A key difference between OWE and TOWE is that OWE does not require the opinion words to tie to any target words in the sentence (i.e., general words to express opinion) while the opinion words in TOWE should be explicitly paired with a given target word. Note that some previous works have also attempted to jointly predict the target and opinion words \citep{Wang:16,Wang:17,Li:17}; however, the target words are still not paired with their corresponding opinion words in these studies \citep{fan2019target}.

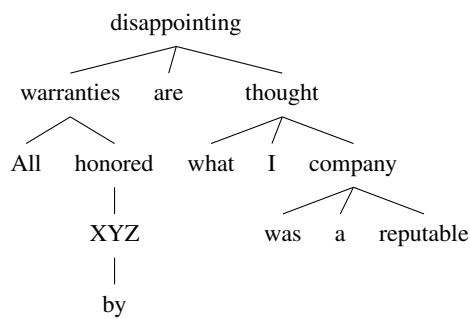
\begin{figure}
\small
\centering
\resizebox{.40\textwidth}{!}{
\begin{forest}
[disappointing
  [warranties
    [All]
    [honored
        [XYZ
            [by]]]]
  [are]
  [thought
    [what]
    [I]
    [company
        [was]
        [a]
        [reputable]]]
]
\end{forest}
}
\caption{\small The dependency tree of the example sentence.} \label{parse}
\end{figure}

%Among the previous works for TOWE, t

The early approach for TOWE has involved the rule-based and lexicon-based methods \cite{hu2004mining,zhuang2006movie} while the recent work has focused on deep learning models for this problem \cite{fan2019target,wu2020latent}. One of the insights from the rule-based methods is that the syntactic structures (i.e., the parsing trees) of the sentences can provide useful information to improve the performance for TOWE \citep{zhuang2006movie}. However, these syntactic structures have not been exploited in the current deep learning models for TOWE \citep{fan2019target,wu2020latent}. Consequently, in this work, we seek to fill in this gap by extracting useful knowledge from the syntactic structures to help the deep learning models learn better representations for TOWE. In particular, based on the dependency parsing trees, we envision two major syntactic information that can be complementarily beneficial for the deep learning models for TOWE, i.e., the syntax-based opinion possibility scores and syntactic word connections for representation learning. First, for the syntax-based possibility scores, our intuition is that the closer words to the target word in the dependency tree of the input sentence tend to have better chance for being the opinion words for the target in TOWE. For instance, in our running example, the opinion word ``\textit{disappointing}" is sequentially far from its target word ``\textit{warranties}". However, in the dependency tree shown in Figure \ref{parse}, ``\textit{disappointing}" is directly connected to ``\textit{warranties}", promoting the distance between ``\textit{disappointing}" and ``\textit{warranties}" (i.e., the length of the connecting path) in the dependency tree as an useful feature for TOWE. Consequently, in this work, we propose to use the distances between the words and the target word in the dependency trees to obtain a score to represent how likely a word is an opinion word for TOWE (called syntax-based possibility scores). These possibility scores would then be introduced into the deep learning models to improve the representation learning for TOWE.

In order to achieve such possibility score incorporation, we propose to employ the representation vectors for the words in the deep learning models to compute a model-based possibility score for each word in the sentence. The model-based possibility scores also aim to quantify the likelihood of being an opinion word for each word in the sentence; however, they are based on the internal representation learning mechanism of the deep learning models for TOWE. To this end, we propose to inject the information from the syntax-based possibility scores into the models for TOWE by enforcing the similarity/consistency between the syntax-based and model-based possibility scores for the words in the sentence. The rationale is to leverage the possibility score consistency to guide the representation learning process of the deep learning models (using the extracted syntactic information) to generate more effective representations for TOWE. In this work, we employ the Ordered-Neuron Long Short-Term Memory Networks (ON-LSTM) \cite{Shen2019ordered} to obtain the model-based possibility scores for the words in the sentences for TOWE. ON-LSTM introduces two additional gates into the original Long Short-Term Memory Network (LSTM) cells that facilitate the computation of the model-based possibility scores via the numbers of active neurons in the hidden vectors for each word.

For the second type of syntactic information in this work, the main motivation is to further improve the representation vector computation for each word by leveraging the dependency connections between the words to infer the effective context words for each word in the sentence. In particular, motivated by our running example, we argue that the effective context words for the representation vector of a current word in TOWE involve the neighboring words of the current word and the target word in the dependency tree. For instance, consider the running example with ``\textit{warranties}" as the target word and ``\textit{reputable}" as the word we need to compute the representation vector. On the one hand, it is important to include the information of the neighboring words of ``\textit{reputable}" (i.e., ``\textit{company}") in the representation so the models can know the context for the current word (e.g., which object ``\textit{reputable}" is modifying). On the other hand, the information about the target word (i.e., ``\textit{warranties}" and possibly its neighboring words) should also be encoded in the representation vector for ``\textit{reputable}" so the models can be aware of the context of the target word and make appropriate comparison in the representation to decide the label (i.e., non-opinion word) for ``\textit{reputable}" in this case. Note that this syntactic connection mechanism allows the models to de-emphasize the context information of ``\textit{I}'' in the representation for ``\textit{reputable}" to improve the representation quality. Consequently, in this work, we propose to formulate these intuitions into an importance score matrix whose cells quantify the contextual importance that a word would contribute to the representation vector of another word, given a target word for TOWE. These importance scores will be conditioned on the distances between the target word and the other words in the dependency tree. Afterward, the score matrix will be consumed by a Graph Convolutional Neural Network (GCN) model \citep{kipf2017semi} to produce the final representation vectors for opinion word prediction.

Finally, in order to further improve the induced representation vectors for TOWE, we introduce a novel inductive bias that seeks to explicitly distinguish the representation vectors of the target-oriented opinion words and those for the other words in the sentence. We conduct extensive experiments to demonstrate the benefits of the proposed model, leading to the state-of-the-art performance for TOWE in several benchmark datasets.

\section{Related Work}

Comparing to the related tasks, TOWE has been relatively less explored in the literature. In particular, the most related task of TOWE is opinion word extraction (OWE) that aims to locate the terms used to express attitude in the sentences \citep{htay2013extracting, shamshurin2012extracting}. A key difference between OWE and TOWE is that OWE does not require the opinion words to tie to any target words in the sentence while the opinion words in TOWE should be explicitly paired with a given target word. Another related task for TOWE is opinion target extraction (OTE) that attempts to identify the target words in the sentences \citep{qiu2011opinion,liu2015fine,poria2016aspect,yin2016unsupervised,xu2018double}. Note that some previous works have also attempted to jointly predict the target and opinion words \citep{qiu2011opinion,liu2013opinion,wang2016recursive,wang2017coupled,li2017deep}; however, the target words are still not paired with their corresponding opinion words in these studies.

%\citep{fan2019target}.

As mentioned in the introduction, among a few previous work on TOWE, the main approaches include the rule-based methods (i.e., based on word distances or syntactic patterns) \citep{zhuang2006movie,hu2004mining} and the recent deep learning models \cite{fan2019target,wu2020latent}. Our model is different from the previous deep learning models as we exploit the syntactic information (i.e., dependency trees) for TOWE with deep learning.

\section{Model}

%The TOWE problem can be formulated as a sequence labeling task. Formally, given a sequence of $N$ words $W=w_1,w_2,...,w_N$ and the target word at index $t$ (i.e., $w_t$), our goal is to assign the label $l_i$ to word $w_i$ to create the label sequence $L=l_1,l_2,...,l_N$. Following previous work, we use BIO tagging schema to encode label $l_i$ (i.e., \textbf{B}eginning, \textbf{I}nside or \textbf{O}utside of the opinion word). 

The TOWE problem can be formulated as a sequence labeling task. Formally, given a sentence $W$ of $N$ words: $W=w_1,w_2,\ldots,w_N$ with $w_t$ as the target word ($1 \le t \le N$), the goal is to assign a label $l_i$ to each word $w_i$ so the label sequence  $L=l_1,l_2,...,l_N$ for $W$ can capture the target-oriented opinion words for $w_t$. Following the previous work \citep{fan2019target}, we use the BIO tagging schema to encode the label $l_i$ for TOWE (i.e., $l_i \in \{B, I, O\}$ for being at the \textbf{B}eginning, \textbf{I}nside or \textbf{O}utside of the opinion words respectively). Our model for TOWE consists of four components that would be described in the following: (i) Sentence Encoding, (ii) Syntax-Model Consistency, (iii) Graph Convolutional Neural Networks, and (iv) Representation Regularization.

%The proposed model consists of the following components: (1) Sentence Encoding: To encode each word into a vector representations (2) Word Importance Encoder: To update word representations to encode the relative importance of the words to the given target (3) Syntax Encoder: To incorporate syntactic structure into the model (4) Syntax-based Regularization: To impose the syntax-based inductive bias to emphasize target-related opinion words (5) Prediction: to assign BIO tags $l_i$ to the words $w_i$ in the sentence $W$. In the following, we provide the details of these components.

\subsection{Sentence Encoding}

In order to represent the input sentence $W$, we encode each word $w_i$ into a real-valued vector $x_i$ based on the concatenation of the two following vectors: (1) the hidden vector of the first wordpiece of $w_i$ from the last layer of the BERT$_{base}$ model \cite{Devlin:19}, and (2) the position embedding for $w_i$. For this vector, we first compute the relative distance $d_i$ from $w_i$ to the target word $w_t$ (i.e., $r_i = i-t$). Afterward, we retrieve the position embedding for $w_i$ by looking up $r_i$ in a position embedding table (initialized randomly). The position embeddings are fine-tuned during training in this work. The resulting vector sequence $X=x_1,x_2,\ldots,x_N$ for $W$ will be then sent to the next computation step.

%We employ a randomly initialized look-up table to compute the embedding of the $i$-th word POS tag. (3) Position embedding: For each word $w_i$, we compute its distance to the target word $w_t$, i.e., $d_i=|i-t|$. Then the embedding of the distance $d_i$ computed from a randomly initialized look-up table is employed in the word representation.  Note that all three types of embedding (i.e., pre-trained word embedding, POS tag embedding and Position embedding) are fine-tuned during training. The word representations $X=x_1,x_2,...,x_N$ is next consumed by the word importance encoder in the model.

\subsection{Syntax-Model Consistency}
\label{sec:importance-encoder}

As presented in the introduction, the goal of this component is to employ the dependency tree of $W$ to obtain the syntax-based opinion possibility scores for the words. These scores would be used to guide the representation learning of the models via the consistency with the model-based possibility scores. In particular, as we consider the closer words to the target word $w_t$ in the dependency tree of $W$ as being more likely to be the target-oriented opinion words, we first compute the distance $d^{syn}_i$ between each word $w_i$ to the target word $w_t$ in the dependency tree (i.e., the number of words along the shortest path between $w_i$ and $w_t$). Afterward, we obtain the syntax-based possibility score $s^{syn}_i$ for $w_i$ based on: $s^{syn}_i = \frac{\exp(-d^{syn}_i)}{\sum_{j=1..N}\exp(-d^{syn}_j)}$.

%In this component, we aim to update the word representations $X$ based on their relative importance to the given target word $w_t$. Firstly, to obtain the relative importance of the word $w_i$ to the target word $w_t$, we exploit dependency tree. More specifically, we first compute the distance between $w_i$ and $w_t$ in the dependency tree, denoted by $d^{syn}_i$. Afterwards, we compute the syntax-based importance score for the $i$ith word as follows:

In order to implement the possibility score consistency, our deep learning model needs to produce $s^{syn}_1,s^{syn}_2,\ldots,s^{syn}_N$ as the model-based possibility scores the words $w_1, w_2, \ldots, w_N$ in $W$ respectively. While the model-based score computation would be explained later, given the model-based scores, the syntax-model consistency for possibility scores would be enforced by introducing the KL divergence $L_{const}$ between the syntax-based and model-based scores into the overall loss function to minimize:
\begin{equation}
\small
    L_{KL} = -\sum_i s^{model}_i \frac{s^{model}_i}{s^{syn}_i}
    \label{eq:klloss}
\end{equation}

%to learn representation vectors for the words in the sentences

As mentioned in the introduction, in this work, we propose to obtain the model-based possibility scores for TOWE using the Ordered-Neuron Long Short-Term Memory Networks (ON-LSTM) \cite{Shen2019ordered}. ON-LSTM is an extension of the popular Long Short-Term Memory Networks (LSTM) that have been used extensively in Natural Language Processing (NLP). Concretely, given the vector sequence $X=x_1,x_2,\ldots,x_N$ as the input, a LSTM layer would produce a sequence of hidden vectors $H=h_1,h_2,\ldots,h_N$ via:
\begin{equation}
\small
\begin{split}
    f_i & = \sigma(W_fx_i+U_fh_{i-1}+b_f) \\
    i_i & = \sigma(W_ix_i+U_ih_{i-1}+b_i) \\
    o_i & = \sigma(W_ox_i+U_oh_{i-1}+b_o) \\
    \hat{c}_i & = tanh(W_cx_i+U_ch_{i-1}+b_c) \\
    c_i & = f_i \circ c_{i-1} + i_i \circ \hat{c}_i, h_i  = o_i \circ tanh(c_i)
\end{split}
\end{equation}
in which $h_0$ is set to zero vector, $\circ$ is the element-wise multiplication, and $f_t$, $i_t$ and $o_t$ are called the forget, input, and output gates respectively.

A major problem with the LSTM cell is that all the dimensions/neurons of the hidden vectors (for the gates) are equally important as these neurons are active/used for all the step/word $i$ in $W$. In other words, the words in $W$ have the same permission to access to all the available neurons in the hidden vectors of the gates in LSTM. This might not be desirable as given a NLP task, the words in a sentence might have different levels of contextual contribution/information for solving the task. It thus suggests a mechanism where the words in the sentences have different access to the neurons in the hidden vectors depending on their informativeness. To this end, ON-LSTM introduces two additional gates $\bar{f}_i$ and $\bar{i}_i$ (the master forget and input gates) into the original LSTM mechanism using the $cummax$ activation function (i.e., $cumax(x) = cumsum(softmax(x))$)\footnote{$cumsum(u_1,u_2,\ldots, u_n) = (u'_1,u'_2,\ldots,u'_n)$ where $u'_i = \sum_{j=1..i}u_j$.}:
\begin{equation}
\small
\begin{split}
    \hat{f}_i & = cummax(W_{\hat{f}}x_i+U_{\hat{f}}h_{i-1}+b_{\hat{f}}) \\
    \hat{i}_i & = 1- cummax(W_{\hat{i}}x_i+U_{\hat{i}}h_{t-1}+b_{\hat{i}}) \\
    \bar{f}_i & = \hat{f}_i \circ (f_i \hat{i}_i + 1 - \hat{i}_i), \bar{i}_i = \hat{i}_i \circ (i_t \hat{f}_i + 1 - \hat{f}_i) \\
    c_i & = \bar{f}_i \circ c_{i-1} + \bar{i}_i \circ \hat{c}_i 
\end{split}
\label{eq:2}
\end{equation}
The benefit of $cummax$ is to introduce a hierarchy over the neurons in the hidden vectors of the master gates so the higher-ranking neurons would be active for more words in the sentence and vice verse (i.e., the activity of the neurons is limited to only a portion of the words in the sentence in this case). In particular, as $cummax$ applies the softmax function on the input vector whose outputs are aggregated over the dimensions, the result of $cummax(x)$ represents the expectation of a binary vector of the form $(0,\ldots,0,1,\ldots,1)$ (i.e., two consecutive segments of 0's and 1's). The 1's segment in this binary vector determines the neurons/dimensions activated for the current step/word $w_i$, thus enabling the different access of the words to the neurons. In ON-LSTM, a word is considered as more informative or important for the task if it has more active neurons (or a larger size for its 1's segment) in the master gates' hidden vectors than the other words in the sentence. As such, ON-LSTM introduces a mechanism to estimate an informativeness score $s^{imp}_i$ for each word $w_i$ in the sentence based on the number of active neurons in the master gates. Following \citep{Shen2019ordered}, we approximate $s^{imp}_i$ via the sum of the weights of the neurons in the master forget gates, i.e., $s^{imp}_i = 1- \sum_{j=1..D} \hat{f}_{ij}$. Here, $D$ is the number of dimensions/neurons in the hidden vectors of the ON-LSTM gates and $\hat{f}_{ij}$ is the weight of the $j$-th dimension for the master forget gate $\hat{f}_i$ at $w_i$.

%ON-LSTM seeks to estimate the informativeness of a word $w_i$ in $W$ via the number of active neurons (i.e., the size of the 1's segment) in the master gates' hidden vectors for the current word. In particular, following \cite{Shen2019ordered}, we approximate the number of active neurons with the sum of the weights of the neurons in the master forget gates, resulting in the estimated informativeness score $s^{imp}_i$ for each word $w_i$ by: $s^{imp}_i = 1- \sum_{j=1..D} \hat{f}_{ij}$. Here, $D$ is the number of dimensions/neurons in the hidden vectors of the ON-LSTM gates and $\hat{f}_{ij}$ is the weight of the $j$-th dimension for the master forget gate $\hat{f}_i$ at $w_i$.

%Consequently, in our TOWE problem, as the target-oriented opinion words are more informative than the other words (i.e., the non-target opinion words and the other words) for the understanding the sentiment of the target words

%An important property in our TOWE problem involves the more informativeness of the target-oriented opinion words over the other words in the sentence (i.e., the non-target opinion words and the other words)

An important property of the target-oriented opinion words in our TOWE problem is that they tend to be more informative than the other words in the sentence (i.e., for understanding the sentiment of the target words). To this end, we propose to compute the model-based opinion possibility scores $s^{model}_i$ for $w_i$ based on the informativeness scores $s^{imp}_i$ from ON-LSTM via: $s^{model}_i = \frac{\exp(s^{imp}_i)}{\sum_{j=1..N}\exp(s^{imp}_j)}$. Consequently, by promoting the syntax-model consistency as in Equation \ref{eq:klloss}, we expect that the syntactic information from the syntax-based possibility scores can directly interfere with the internal computation/structure of the ON-LSTM cell (via the neurons of the master gates) to potentially produce better representation vectors for TOWE. For convenience, we also use $H=h_1, h_2, \ldots, h_N$ to denote the hidden vectors returned by running ON-LSTM over the input sequence vector $X$ in the following.

\subsection{Graph Convolutional Networks}
\label{sec:syntax-encoder}

%This component seeks to employ the connections between the words in the dependency tree of $W$ to extract effective context words for better representation learning in TOWE. 

This component seeks to extract effective context words to further improve the representation vectors $H$ for the words in $W$ based on the dependency connections between the words for TOWE. As discussed in the introduction, given the current word $w_i \in W$, there are two groups of important context words in $W$ that should be explicitly encoded in the representation vector for $w_i$ to enable effective opinion word prediction: (i) the neighboring words of $w_i$, and (ii) the neighboring words of the target word $w_t$ in the dependency tree (i.e., these words should receive higher weights than the others in the representation computation for $w_i$). Consequently, in order to capture such important context words for all the words in the sentence for TOWE, we propose to obtain two importance score matrices of size $N \times N$ for which the scores at cells $(i,j)$ are expected to weight the importance of the contextual information from $w_j$ with respect to the representation vector computation for $w_i$ in $W$. In particular, one score matrix would be used to capture the syntactic neighboring words of the current words (i.e., $w_i$) while the other score matrix would be reserved for the neighboring words of the target word $w_t$. These two matrices would then be combined and consumed by a GCN model \citep{kipf2017semi} for representation learning. 

Specifically, for the syntactic neighbors of the current words, following the previous GCN models for NLP \citep{Marcheggiani:17,Nguyen:18,Veyseh:19b}, we directly use the adjacency binary matrix $A^d = \{a^d_{i,j}\}_{i,j=1..N}$ of the dependency tree for $W$ as the importance score matrix for this group of words. Note that $a^d_{i,j}$ is only set to 1 if $w_i$ is directly connected to $w_j$ in the dependency tree or $i=j$ in this case. In the next step for the neighboring words of the target word $w_t$, as we expect the closer words to the target word $w_t$ to have larger contributions for the representation vectors of the words in $W$ for TOWE, we propose to use the syntactic distances (to the target word) $d^{syn}_i$ and $d^{syn}_j$ of $w_i$ and $w_j$ as the features to learn the importance score matrix $A^t = \{a^t_{i,j}\}_{i,j=1..N}$ for the words in this case. In particular, $a^t_{i,j}$ would be computed by: $a^t_{i,j} = \sigma(FF([d^{syn}_i, d^{syn}_j, d^{syn}_i + d^{syn}_j, |d^{syn}_i - d^{syn}_j|, d^{syn}_i * d^{syn}_j]))$ where $FF$ is a feed-forward network to convert a vector input with five dimensions into a scalar score and $\sigma$ is the {\it sigmoid} function. Given the importance score matrices $A^d$ and $A^t$, we seek to integrate them into a single importance score matrix $A$ to simultaneously capture the two groups of important context words for representation learning in TOWE via the weighted sum: $A = \gamma A^d + (1-\gamma) A^t = \{a_{i,j}\}_{i,j=1..N}$ where $\gamma$ is a trade-off parameter\footnote{Note that we tried to directly learn $A$ from the available information from $A^d$ and $A^t$ (i.e., $a_{i,j} = \sigma(FF([a^d_{i,j}, d^{syn}_i, d^{syn}_j, d^{syn}_i + d^{syn}_j, |d^{syn}_i - d^{syn}_j|, d^{syn}_i * d^{syn}_j]))$). However, the performance of this model was worse than the linear combination of $A^d$ and $A^t$ in our experiments.}.

In the next step for this component, we run a GCN model over the ON-LSTM hidden vectors $H$ to learn more abstract representation vectors for the words in $W$. This step will leverage $A$ as the adjacency matrix to enrich the representation vector for each word $w_i$ with the information from its effective context words (i.e., the syntactic neighboring words of $w_i$ and $w_t$), potentially improving the opinion word prediction for $w_i$. In particular, the GCN model in this work involves several layers (i.e., $G$ layers in our case). The representation vector $\bar{h}^k_i$ for the word $w_i$ at the $k$-the layer of the GCN model would be computed by:
\begin{equation}
\small
    \begin{split}
             %\hat{h}^k_i & = ReLU(W_k \bar{h}^{k-1}_i + b_k) \\
             \bar{h}^k_i & = ReLU\left(\frac{\Sigma_{j=1..N} a_{i,j}(W_k \bar{h}^{k-1}_j + b_k)}{\sum_{j=1..N} a_{i,j}}\right)
    \end{split}
\end{equation}
where $W_k$ and $b_k$ are the weight matrix and bias for the $k$-th GCN layer. The input vector $h^0_i$ for GCN is set to the hidden vector $h_i$ from ON-LSTM (i.e., $h^0_i = h_i$) for all $i$ in this case. For convenience, we denote $\bar{h}_i$ as the hidden vector for $w_i$ in the last layer of GCN (i.e., $\bar{h}_i = \bar{h}^G_i$ for all $1 \le i \le N$). We also write $\bar{h}_1,\bar{h}_2,\ldots,\bar{h}_N = GCN(H, A)$ to indicate that $\bar{h}_1,\bar{h}_2,\ldots,\bar{h}_N$ are the hidden vectors in the last layer of the GCN model run over the input $H$ and the adjacency matrix $A$ for simplicity.

Finally, given the syntax-enriched representation vectors $h_i$ from ON-LSTM and $\bar{h}_i$ from the last layer of GCN, we form the vector $V_i = [h_i, \bar{h}_i]$ to serve as the feature to perform opinion word prediction for $w_i$. In particular, $V_i$ would be sent to a two-layer feed-forward network with the softmax function in the end to produce a probability distribution $P(.|W,t,i)$ over the possible opinion labels for $w_i$ (i.e., B, I, and O). The negative log-likelihood function $L_{pred}$ would then be used as the objective function to train the overall model: $L_{pred} = -\sum_{i=1}^N P(l_i|W,t,i)$.

\subsection{Representation Regularization}
\label{sec:reg}
There are three groups of words in the input sentence $W$ for our TOWE problem, i.e., the target word $w_t$, the target-oriented opinion words (i.e., the words we want to identify) (called $W^{opinion}$), and the other words (called $W^{other}$). After the input sentence $W$ has been processed by several abstraction layers (i.e., ON-LSTM and GCN), we expect that the resulting representation vectors for the target word and the target-oriented opinion words would capture the sentiment polarity information for the target word while the representation vectors for the other words might encode some other context information in $W$. We thus argue that the representation vector for the target word should be more similar to the representations for the words in $W^{opinion}$ (in term of the sentiment polarity) than those for $W^{other}$. To this end, we introduce an explicit loss term to encourage such representation distinction between these groups of words to potentially promote better representation vectors for TOWE. In particular, let $R^{tar}$, $R^{opn}$, and $R^{oth}$ be some representation vectors for the target word $w_t$, the target-oriented opinion words (i.e., $W^{opinion}$), and the other words (i.e., $W^{other}$) in $W$. The loss term for the representation distinction based on our intuition (i.e., to encourage $R^{tar}$ to be more similar to $R^{opn}$ than $R^{oth}$) can be captured via the following triplet loss for minimization:
\begin{equation}
\label{eq:reg-loss}
\small
    L_{reg} = 1 - cosine(R^{tar}, R^{opn}) + cosine(R^{tar}, R^{oth})
\end{equation}
%where $cos$ is the cosine function.

In this work, the representation vector for the target word is simply taken from last GCN layer, i.e., $R^{tar} = \bar{h}_t$. However, as $W^{opinion}$ and $W^{other}$ might involve sets of words, we need to aggregate the representation vectors for the individual words in these sets to produce the single representation vectors $R^{opn}$ and $R^{oth}$. The simple and popular aggregation method in this case involves performing the max-pooling operation over the representation vectors (i.e., from GCN) for the individual words in each set (i.e., our baseline). However, this approach ignores the structures/orders of the individual words in $W^{opinion}$ and $W^{other}$, and fails to recognize the target word for better customized representation for regularization. To this end, we propose to preserve the syntactic structures among the words in $W^{opinion}$ and $W^{other}$ in the representation computation for regularization for these sets. This is done by generating the target-oriented pruned trees from the original dependency tree for $W$ that are customized for the words in $W^{opinion}$ and $W^{other}$. These pruned trees would then be consumed by the GCN model in the previous section to produce the representation vectors for $W^{opinion}$ and $W^{other}$ in this part. In particular, we obtain the pruned tree for the target-oriented opinion words $W^{opinion}$ by forming the adjacency matrix $A^{opinion} = \{a^{opinion}_{i,j}\}_{i,j=1..N}$ where $a^{opinion}_{i,j} = a_{i,j}$ if both $w_i$ and $w_j$ belong to some shortest dependency paths between $w_t$ and some words in $W^{opinion}$, and 0 otherwise. This helps to maintain the syntactic structures of the words in $W^{opinion}$ and also introduce the target word $w_t$ as the center of the pruned tree for representation learning. We apply the similar procedure to obtain the adjacency matrix $A^{other} = \{a^{other}_{i,j}\}_{i,j=1..N}$ for the pruned tree for $W^{other}$. Given the two adjacency matrices for the pruned trees, the GCN model in the previous section is run over the ON-LSTM vectors $H$, resulting in two sequences of hidden vectors for $W^{opinion}$ and $W^{other}$, i.e., $h'_1,h'_2,\ldots,h'_N = GCN(H, A^{opinion})$ and $h''_1,h''_2,\ldots,h''_N = GCN(H, A^{other})$. Afterward, we compute the representation vectors $R^{opn}$ and $R^{oth}$ for the sets $W^{opinion}$ and $W^{other}$ by retrieving the hidden vectors for the target word returned by the GCN model with the corresponding adjacency matrices, i.e., $R^{opn} = h'_t$ and $R^{oth} = h''_t$. Note that the application of GCN over the pruned trees and the ON-LSTM vectors makes $R^{opn}$ and $R^{oth}$ more comparable with $R^{tar}$ in our case. This completes the description for the representation regularizer in this work. The overall loss function in this work would be: $L = L_{pred} + \alpha L_{KL} + \beta L_{reg}$ where $\alpha$ and $\beta$ are the trade-off parameters.

\section{Experiments}
\subsection{Datasets \& Parameters}
%We use four benchmark datasets presented in \cite{fan2019target} to evaluate the effectiveness of the proposed TOWE model. These datasets contain reviews for restaurants (i.e., datasets 14res, 15res and 16res) or laptops, (i.e., dataset 14lap). They are created from the widely used ABSA datasets from the SemEval challenges (i.e., SemEval 2014 Task4 (14res and 14lap), SemEval 2015 task 12 (15res) and SemEval 2016 task 5 (16res)). Table \ref{tab:stat} shows the statistics of the datasets. In these datasets, a sentence might have multiple targets, therefore, these sentences are repeated as many as targets mentioned in them. Note that for each target the opinion word is explicitly expressed in the sentence. To measure the performance of our model on these datasets, following previous work \cite{fan2019target,wu2020latent}, we use precision, recall and F1 score for the opinion words in the sentence. More specifically, a predicted opinion word is regarded correct if the beginning and the end of the prediction matches with the golden data in the datasets.

We use four benchmark datasets presented in \citep{fan2019target} to evaluate the effectiveness of the proposed TOWE model. These datasets contain reviews for restaurants (i.e., the datasets {\bf 14res}, {\bf 15res} and {\bf 16res}) and laptops, (i.e., the dataset {\bf 14lap}). They are created from the widely used ABSA datasets from the SemEval challenges (i.e., SemEval 2014 Task 4 (14res and 14lap), SemEval 2015 Task 12 (15res) and SemEval 2016 Task 5 (16res)). Each example in these datasets involves a target word in a sentence where the opinion words have been manually annotated. 

%Following the prior work \citep{fan2019target}, we use Precision, Recall and F1 as the performance measure in this work.

%As a sentence in these datasets might involve multiple target words, the sentences will be repeated as many as the target words in them and each example in the datasets will correspond to a target word in the sentences

%(i.e., an opinion word is considered correct if its boundary matches one opinion word in the golden data).

%(i.e., the same samples for the development data as \citep{fan2019target} for a fair comparison)

As none of the datasets provides the development data, for each dataset, we sample 20\% of the training instances for the development sets. Note that we use the same samples for the development data as in \citep{fan2019target} to achieve a fair comparison. We use the 14res development set for hyper-parameter fine-tuning, leading to the following values for the proposed model (used for all the datasets): 30 dimensions for the position embeddings, 200 dimensions for the layers of the feed-forward networks and GCN (with $G=2$ layers), 300 hidden units for one layer of ON-LSTM, 0.2 for $\gamma$ in $A$, and 0.1 for the parameters $\alpha$ and $\beta$.

%A reproducibility checklist is shown in Appendix \ref{app:repo}.

%the importance score matrix 

%in the overall loss function

%As none of the datasets provides the development data, for each dataset, we randomly choose 10\% of the training instances for development set and train the model on the remaining portion of the train set. We use the development set of each dataset to fine-tuen the hyper-parameters of the model. Through this process, the following values are found: 746 hidden dimension from the last layer of the BERT$_{base}$ model for pre-trained word embedding; 30 dimensions for POS tag and position embedding; 200 dimension for all feed forward neural nets and graph convolution network; 300 dimension of hidden states for 2 layers of ON-LSTM; 0.2 for syntax encoder trade-off parameter (i.e., $\alpha$) and 0.1 for loss function trade-off parameters (i.e., $\beta$ and $\gamma$).

%\begin{table}[t!]
%%\small
%%\addtolength{\abovecaptionskip}{-4.0mm}
%%\addtolength{\belowcaptionskip}{-3.mm}
%\begin{center}
%% \resizebox{.8\textwidth}{!}{
%\begin{tabular}{l|cc|cc}
%  & \multicolumn{2}{c}{\#sentences} & \multicolumn{2}{c}{\#targets} \\
%  & Train & Test & Train & Test \\ \hline
%  14res & 1627 & 500 & 2643 & 865 \\ \hline
%  14lap & 1158 & 343 & 1634 & 482 \\ \hline
%  15res & 754 & 325 & 1076 & 436 \\ \hline
%  16res & 1079 & 329 & 1512 & 457
%\end{tabular}
% }
%\end{center}
%\caption{\label{tab:stat} Statistics of the datasets. A sentence might contain multiple targets and the number of targets is equal to the number of instances.
%  }
%\end{table}

\subsection{Comparing to the State of the Art}

\begin{table*}[t!]
%\small
%\addtolength{\abovecaptionskip}{-4.0mm}
%\addtolength{\belowcaptionskip}{-3.mm}
\begin{center}
\resizebox{.95\textwidth}{!}{
\begin{tabular}{l|ccc|ccc|ccc|ccc}
  & \multicolumn{3}{c}{14res} & \multicolumn{3}{c}{14lap} & \multicolumn{3}{c}{15res} & \multicolumn{3}{c}{16res} \\
  Model & P & R & F1 & P & R & F1 & P & R & F1 & P & R & F1 \\
  \hline
  Distance-rule \shortcite{hu2004mining} & 58.39 & 43.59 & 49.92 & 50.13 & 33.86 & 40.42 & 54.12 & 39.96 & 45.97 & 61.90 & 44.57 & 51.83 \\
  Dependency-rule \shortcite{zhuang2006movie} & 64.57 & 52.72 & 58.04 & 45.09 & 31.57 & 37.14 & 65.49 & 48.88 & 55.98 & 76.03 & 56.19 & 64.62 \\ 
     LSTM \shortcite{liu2015fine} & 52.64 & 65.47 & 58.34 & 55.71 & 57.53 & 56.52 & 57.27 & 60.69 & 58.93 & 62.46 & 68.72 & 65.33 \\
    BiLSTM \shortcite{liu2015fine}  & 58.34 & 61.73 & 59.95 & 64.52 & 61.45 &  62.71 & 60.46 & 63.65 & 62.00 & 68.68 & 70.51 & 69.57 \\
 Pipeline \shortcite{fan2019target} & 77.72 & 62.33 & 69.18 & 72.58 & 56.97 & 63.83 & 74.75 & 60.65 & 66.97 & 81.46 & 67.81 & 74.01 \\
    TC-BiLSTM \shortcite{fan2019target} & 67.65 & 67.67 & 67.61 & 62.45 &  60.14 & 61.21 & 66.06 & 60.16 & 62.94 & 73.46 & 72.88 & 73.10 \\
     IOG \shortcite{fan2019target} & 82.85 & 77.38 & 80.02 & 73.24 & 69.63 & 71.35 & 76.06 & 70.71 & 73.25 & 82.25 & 78.51 & 81.69 \\
     LOTN \cite{wu2020latent} & 84.00 & 80.52 & 82.21 & 77.08 & 67.62 & 72.02 & 76.61 & 70.29 & 73.29 & 86.57 & 80.89 & 83.62 \\
        \hline
        \textbf{ONG (Ours)} & 83.23 & 81.46 & {\bf 82.33} & 73.87 & 77.78 & {\bf 75.77} & 76.63 & 81.14 & {\bf 78.81} & 87.72 & 84.38 & {\bf 86.01} \\
\end{tabular}
}
\end{center}
\caption{\label{tab:results} Test set performance (i.e., Precision (P), Recall (R) and F1 scores) of the models.
  }
\end{table*}

We compare the TOWE model in this work (called {\bf ONG} for ON-LSTM and GCN) with the recent models in \cite{fan2019target,wu2020latent} and their baselines. More specifically, the following baselines are considered in our experiments:

%\begin{itemize}
%    \item 
    
    1. \textbf{Rule-based}: These baselines employ predefined patterns to extract the opinion-target pairs that could be either {\bf dependency-based} \citep{zhuang2006movie} or {\bf distance-based} \citep{hu2004mining}.
%    \item 
    
    2. \textbf{Sequence-based Deep Learning}: These approaches apply some deep learning model over the input sentences following the sequential order of the words to predict the opinion words (i.e., {\bf LSTM}/{\bf BiLSTM} \citep{liu2015fine}, {\bf TC-BiLSTM} \citep{fan2019target} and {\bf IOG} \citep{fan2019target}).
%    \item 
    
    3. \textbf{Pipeline with Deep Learning}: This method utilizes a recurrent neural network to predict the opinion words. The distance-based rules are then introduced to select the target-oriented opinion words (i.e., {\bf Pipeline}) \cite{fan2019target}.
%    \item 
    
    4. \textbf{Multitask Learning}: These methods seek to jointly solve TOWE and another related task (i.e., sentiment classification). In particular, the {\bf LOTN} model in \cite{wu2020latent} uses a pre-trained SA model to obtain an auxiliary label for each word in the sentence using distance-based rules. A bidirectional LSTM model is then trained to make prediction for both TOWE and the auxiliary labels\footnote{Note that \citep{peng2019knowing} also proposes a related model for TOWE based on multitask deep learning. However, the models in this work actually predict general opinion words that are not necessary tied to any target word. As we focus on target-oriented opinion words, the models in \citep{peng2019knowing} are not comparable with us.}.

Table \ref{tab:results} shows the performance of the models on the test sets of the four datasets. It is clear from the table that the proposed ONG model outperforms all the other baseline methods in this work. The performance gap between ONG and the other models are large and significant (with $p < 0.01$) over all the four benchmark datasets (except for LOTN on 14res), clearly testifying to the effectiveness of the proposed model for TOWE. Among different factors, we attribute this better performance of ONG to the use of syntactic information (i.e., the dependency trees) to guide the representation learning of the models (i.e., with ON-LSTM and GCN) that is not considered in the previous deep learning models for TOWE.

\subsection{Model Analysis and Ablation Study}

There are three main components in the proposed ONG model, including the ON-LSTM component, the GCN component and the representation regularization component. This section studies different variations and ablated versions of such components to highlight their importance for ONG.

%effectiveness of the designed model.

%for TOWE in this work.

{\bf ON-LSTM}: First, we evaluate the following variations for the ON-LSTM component: (i) {\bf ONG - KL}: this model is similar to ONG, except that the syntax-model consistency loss based on KL $L_{KL}$ is not included in the overall loss function, (ii) {\bf ONG - ON-LSTM}: this model completely removes the ON-LSTM component in ONG (so the KL-based syntax-model consistency loss is not used and the input vector sequence $X$ is directly sent to the GCN model), and (iii) {\bf ONG\_wLSTM}: this model replaces the ON-LSTM model with the traditional LSTM model in ONG (so the syntax-model consistency loss is also not employed in this case as LSTM does not support the neuron hierarchy for model-based possibility scores). The performance for these models on the test sets (i.e., F1 scores) are presented in Table \ref{tab:on-lstm}.

\begin{table}[ht]
%\small
    \centering
    \resizebox{.44\textwidth}{!}{
    \begin{tabular}{l|c|c|c|c}
        Model & 14res & 14lap & 15res & 16res \\ \hline
        ONG & 82.33 & 75.77 & 78.81 & 86.01 \\ \hline
        ONG - KL &  80.91 & 73.34 & 76.21 & 83.78 \\
        ONG - ON-LSTM & 78.99 & 70.28 & 71.39 & 81.13 \\
        ONG\_wLSTM &  81.03 & 73.98 & 74.43 & 82.81 \\
    \end{tabular}
    }
    \caption{Performance of the ON-LSTM's variations.}
    \label{tab:on-lstm}
\end{table}
As we can see from the table, the syntax-model consistency loss with KL divergence is important for ONG as removing it would significantly hurt the model's performance on different datasets. The model also becomes significantly worse when the ON-LSTM component is eliminated or replaced by the LSTM model. These evidences altogether confirm the benefits of the ON-LSTM model with the syntax-model consistency proposed in this work.

{\bf GCN Structures}: There are two types of importance score matrices in the GCN model, i.e., the adjacency binary matrices $A^d$ for the syntactic neighbors of the current words and $A^t$ for the syntactic neighbors of the target word. This part evaluates the effectiveness of these score matrices by removing each of them from the GCN model, leading to the two ablated models {\bf ONG - $A^d$} and {\bf ONG - $A^t$} for evaluation. Table \ref{tab:gcn-structure} provides the performance on the test sets for these models (i.e., F1 scores). It is clear from the table that the absence of any importance score matrices (i.e., $A^d$ or $A^t$) would decrease the performance over all the four datasets and both matrices are necessary for ONG to achieve its highest performance.
\begin{table}[ht]
%\small
    \centering
    \resizebox{.4\textwidth}{!}{
    \begin{tabular}{l|c|c|c|c}
        Model & 14res & 14lap & 15res & 16res \\ \hline
        ONG & 82.33 & 75.77 & 78.81 & 86.01 \\ \hline
        ONG - $A^d$ & 80.98 & 73.05 & 75.51 & 83.72 \\
        ONG - $A^t$ & 81.23 & 74.18 & 76.32 & 85.20 \\
    \end{tabular}
    }
    \caption{Ablation study on the GCN structures.}
    \label{tab:gcn-structure}
\end{table}

{\bf GCN and Representation Regularization}: As the representation regularization component relies on the GCN model to obtain the representation vectors, we jointly perform analysis for the GCN and representation regularization components in this part. In particular, we consider the following variations for these two components: (i) {\bf ONG - REG}: this model is similar to ONG except that the representation regularization loss $L_{reg}$ is not applied in the overall loss function, (ii) {\bf ONG\_REG\_wMP-GCN}: this is also similar to ONG; however, it does not apply the GCN model to compute the representation vectors $R^{opn}$ and $R^{oth}$ for regularization. Instead, it uses the simple max-pooling operation over the GCN-produced vectors $\bar{h}_1,\bar{h}_2,\ldots,\bar{h}_N$ of the target-oriented words $W^{opinion}$ and the other words $W^{other}$ for $R^{opn}$ and $R^{oth}$: $R^{opn} = max\_pool(\bar{h}_i|w_i \in W^{opinion})$ and $R^{oth} = max\_pool(\bar{h}_i|w_i \in W^{other})$, (iii) {\bf ONG - GCN}: this model eliminates the GCN model from ONG, but still applies the representation regularization over the representation vectors obtained from the ON-LSTM hidden vectors. In particular, the ON-LSTM hidden vectors $H=h_1,h_2,\ldots,h_N$ would be employed for both opinion word prediction (i.e., $V = [h_i]$ only) and the computation of $R^{target}$, $R^{opn}$ and $R^{oth}$ for representation regularization with max-pooling (i.e., $R^{target} = h_t$, $R^{opn} = max\_pool(h_i|w_i \in W^{opinion})$ and $R^{oth} = max\_pool(h_i|w_i \in W^{other})$) in this case, and (iv) {\bf ONG - GCN - REG}: this model completely excludes both the GCN and the representation regularization models from ONG (so the ON-LSTM hidden vectors $H=h_1,h_2,\ldots,h_N$ are used directly for opinion word prediction (i.e., $V = [h_i]$ as in ONG - GCN) and the regularization loss $L_{reg}$ is not included in the overall loss function). Table \ref{tab:gcn-reg} shows the performance of the models on the test datasets (i.e., F1 scores).
\begin{table}[ht]
%\small
    \centering
    \resizebox{.46\textwidth}{!}{
    \begin{tabular}{l|c|c|c|c}
        Model & 14res & 14lap & 15res & 16res \\ \hline
        ONG & 82.33 & 75.77 & 78.81 & 86.01 \\ \hline
        ONG - REG & 80.88 & 73.89 & 75.92 & 84.03 \\
        ONG\_REG\_wMP-GCN & 80.72 & 72.44  & 74.28 & 84.29 \\
        ONG - GCN & 81.01 & 70.88 & 72.98 & 82.58 \\
        ONG - GCN - REG & 79.23 & 71.04 & 72.53 & 82.13 \\
    \end{tabular}
    }
    \caption{Performance of the variations of the GCN and representation regularization components.}
    \label{tab:gcn-reg}
\end{table}

There are several important observations from this table. First, as ONG - REG is significantly worse than the full model ONG over different datasets, it demonstrates the benefits of the representation regularization component in this work. Second, the better performance of ONG over ONG\_REG\_wMP-GCN (also over all the four datasets) highlights the advantages of the GCN-based representation vectors $R^{opn}$ and $R^{oth}$ over the max-pooled vectors for representation regularization. We attribute this to the ability of ONG to exploit the syntactic structures among the words in $W^{opinion}$ and $W^{other}$ for regularization in this case. Finally, we also see that the GCN model is crucial for the operation of the proposed model as removing it significantly degrades ONG's performance (whether the representation regularization is used (i.e., in ONG - GCN) or not (i.e., in ONG - GCN - REG). The performance become the worst when both the GCN and the regularization components are eliminated in ONG, eventually confirming the effectiveness of our model for TOWE in this work.

%Finally, we provide an additional analysis for the representation regularization in Appendix \ref{app:rr}.

%\begin{table*}[t!]
%%\small
%%\addtolength{\abovecaptionskip}{-4.0mm}
%%\addtolength{\belowcaptionskip}{-3.mm}
%\begin{center}
%\resizebox{.95\textwidth}{!}{
%\begin{tabular}{c|cc|cc|cc|cc}
%  & \multicolumn{2}{c}{14res} & \multicolumn{2}{c}{14lap} & \multicolumn{2}{c}{15res} & \multicolumn{2}{c}{16res} \\
%  Distance & Graph Pooling & Max Pooling & Graph Pooling & Max Pooling & Graph Pooling & Max Pooling & Graph Pooling & Max Pooling \\
%  \hline
%  1 & 83.22 & 79.94 & 76.91 & 75.21 & 79.92 & 74.29 & 86.52 & 83.33 \\
%  2 & 83.18 & 78.43 & 75.03 & 73.12 & 78.04 & 73.33 & 87.31 & 83.27 \\ 
 %    3 & 81.56 & 75.41 & 74.21 & 70.69 & 77.71 & 70.91 & 84.77 & 78.63 \\
 %   4$\le$  & 80.97 & 73.77 & 73.92 & 66.23 & 76.98 & 68.88 & 84.05 & 77.13 \\
%\end{tabular}
%}
%\end{center}
%\caption{\label{tab:reg-Analysis} Performance of our model and Max Pooling regularization baseline for sentences with difference distance between the target and the opinion word
%  }
%\end{table*}

\begin{table}[t!]
%\small
%\addtolength{\abovecaptionskip}{-4.0mm}
%\addtolength{\belowcaptionskip}{-3.mm}
\begin{center}
\resizebox{.48\textwidth}{!}{
\begin{tabular}{c|c|c|c|c}
  & \multicolumn{2}{c|}{14res} & \multicolumn{2}{c}{14lap} \\ \cline{2-5}
  Distance & ONG & ONG\_REG & ONG & ONG\_REG \\
  & & \_wMP-GCN & & \_wMP-GCN \\
  \hline
  1 & 83.22 & 79.94 & 76.91 & 75.21 \\
  2 & 83.18 & 78.43 & 75.03 & 73.12  \\ 
     3 & 81.56 & 75.41 & 74.21 & 70.69 \\
    $>$3  & 80.97 & 73.77 & 73.92 & 66.23 \\ \hline
& \multicolumn{2}{c|}{15res} & \multicolumn{2}{c}{16res} \\ \cline{2-5}
%Distance & ONG & ONG\_REG\_wMP-GCN & ONG & ONG\_REG\_wMP-GCN \\ \hline
Distance & ONG & ONG\_REG & ONG & ONG\_REG \\
  & & \_wMP-GCN & & \_wMP-GCN \\ \hline
 1 & 79.92 & 74.29 & 86.52 & 83.33 \\
 2 & 78.04 & 73.33 & 87.31 & 83.27 \\
 3  & 77.71 & 70.91 & 84.77 & 78.63 \\
 $>$3 & 76.98 & 68.88 & 84.05 & 77.13\\
\end{tabular}
}
\end{center}
\caption{\label{tab:reg-Analysis} The performance (i.e., F1 scores) of ONG and ONG\_REG\_wMP-GCN on the four data folds of the development sets for 14res, 14lap, 15res, and 16res. The data folds are based on the target-opinion distances of the examples (called Distance in this table).}
\end{table}

%In section \ref{sec:reg}, we argue that the representation vector for the target word should be more similar to target-oriented opinion words' (i.e., $W^{opinion}$) than those for the other words in the sentence (i.e., $W^{other}$). In order to enforce this similarity constraint in the model, we introduce the triple loss for representation regularization in Equation \ref{eq:reg-loss} where $R^{opn}$ and $R^{oth}$ stand for the representation vectors for the target-oriented opinion words $W^{opinion}$ and the other words $W^{other}$ (respectively). In this work, we propose to compute the representation vectors $R^{opn}$ and $R^{oth}$ by running a GCN model over the target-oriented pruned dependency trees that are customized for the words in $W^{opinion}$ and $W^{other}$ (respectively). The rationale is to exploit the dependency structures among the words in $W^{opinion}$ and $W^{other}$ to improve the representation vectors $R^{opn}$ and $R^{oth}$ for the regularization purpose. Previous experiments have shown that this graph-based representation computation for $R^{opn}$ and $R^{oth}$ is significantly better than directly performing max-pooling over the representation vectors for the words in $W^{opinion}$ and $W^{other}$, thereby demonstrating the benefit of the dependency structures for $R^{opn}$ and $R^{oth}$.

\textbf{Regularization Analysis}: This section aims to further investigate the effect of the dependency structures $R^{opn}$ and $R^{oth}$ (i.e., among the words in $W^{opinion}$ and $W^{other}$) to gain a better insight into their importance for the representation regularization in this work. Concretely, we again compare the performance of the full proposed model ONG (with the graph-based representations for $R^{opn}$ and $R^{oth}$) and the baseline model ONG\_REG\_wMP-GCN (with the direct max-pooling over the word representations, i.e., $R^{opn} = max\_pool(\bar{h}_i|w_i \in W^{opinion})$ and $R^{oth} = max\_pool(\bar{h}_i|w_i \in W^{other})$). However, in this analysis, we further divide the sentences in the development sets into four folds and observe the models' performance on those fold. As such, for each sentence, we rely on the longest distance between the target word and some target-oriented opinion word in $W^{opinion}$ in the dependency tree to perform this data split (called the target-opinion distance). In particular, the four data folds for the development sets (of each dataset) correspond to the sentences with the target-opinion distances of 1, 2, 3 or greater than 3. Intuitively, the higher target-opinion distances amount to more complicated dependency structures among the target-oriented opinion word in $W^{opinion}$ (as more words are involved in the structures). The four data folds are thus ordered in the increasing complexity levels of the dependency structures in $W^{opinion}$.

Table \ref{tab:reg-Analysis} presents the performance of the models on the four data folds for the development sets of the datasets in this work. First, it is clear from the table that ONG significantly outperforms the baseline model ONG\_REG\_wMP-GCN over all the datasets and structure complexity levels of $W^{opinion}$. Second, we see that as the structure complexity (i.e., the target-opinion distance) increases, the performance of both ONG and ONG\_REG\_wMP-GCN decreases, demonstrating the more challenges presented by the sentences with more complicated dependency structures in $W^{opinion}$ for TOWE. However, comparing ONG and ONG\_REG\_wMP-GCN, we find that ONG's performance decreases slower than those for ONG\_REG\_wMP-GCN when the target-opinion distance increases (for all the four datasets considered in this work). This implies that the complicated dependency structures in $W^{opinion}$ have more detrimental effect on the model's performance for ONG\_REG\_wMP-GCN than those for ONG, leading to the larger performance gaps between ONG and ONG\_REG\_wMP-GCN. Overall, these evidences suggest that the sentences with complicated dependency structures for the words in $W^{opinion}$ are more challenging for the TOWE models and modeling such dependency structures to compute the representation vectors $R^{opn}$ and $R^{oth}$ for regularization (as in ONG) can help the models to better perform on these cases.

\section{Conclusion}

We propose a novel deep learning model for TOWE that seeks to incorporate the syntactic structures of the sentences into the model computation. Two types of syntactic information are introduced in this work, i.e., the syntax-based possibility scores for words (integrated with the ON-LSTM model) and the syntactic connections between the words (applied with the GCN model with novel adjacency matrices). We also present a novel inductive bias to improve the model, leveraging the representation distinction between the words in TOWE. Comprehensive analysis is done to demonstrate the effectiveness of the proposed model over four datasets.

\section*{Acknowledgement}

%This research has been supported in part by Vingroup Innovation Foundation (VINIF) in project code VINIF.2019.DA18 and Adobe Research Gift. This research is also based upon work supported in part by the Office of the Director of National Intelligence (ODNI), Intelligence Advanced Research Projects Activity (IARPA), via IARPA Contract No. 2019-19051600006 under the Better Extraction from Text Towards Enhanced Retrieval (BETTER) Program. The views and conclusions contained herein are those of the authors and should not be interpreted as necessarily representing the official policies, either expressed or implied, of ODNI, IARPA, the Department of Defense, or the U.S. Government. The U.S. Government is authorized to reproduce and distribute reprints for governmental purposes notwithstanding any copyright annotation therein. This document does not contain technology or technical data controlled under either the U.S. International Traffic in Arms Regulations or the U.S. Export Administration Regulations.

This research is based upon work supported in part by the Office of the Director of National Intelligence (ODNI), Intelligence Advanced Research Projects Activity (IARPA), via IARPA Contract No. 2019-19051600006 under the Better Extraction from Text Towards Enhanced Retrieval (BETTER) Program. The views and conclusions contained herein are those of the authors and should not be interpreted as necessarily representing the official policies, either expressed or implied, of ODNI, IARPA, the Department of Defense, or the U.S. Government. The U.S. Government is authorized to reproduce and distribute reprints for governmental purposes notwithstanding any copyright annotation therein. This document does not contain technology or technical data controlled under either the U.S. International Traffic in Arms Regulations or the U.S. Export Administration Regulations.

\bibliography{emnlp2020}
\bibliographystyle{acl_natbib}

\clearpage

\appendix

\end{document}